\documentclass[conference]{IEEEtran}

\usepackage{cite}
\usepackage{amsmath,amssymb,amsfonts}
\usepackage{algorithmic}
\usepackage{graphicx}
\usepackage{textcomp}
\usepackage{xcolor}
\usepackage{booktabs}
\usepackage{multirow}
\usepackage{hyperref}
\usepackage{tikz}
\usetikzlibrary{shapes.geometric, arrows.meta, positioning, fit, backgrounds}

\def\BibTeX{{\rm B\kern-.05em{\sc i\kern-.025em b}\kern-.08em
    T\kern-.1667em\lower.7ex\hbox{E}\kern-.125emX}}

\begin{document}

\title{Energy-Efficient On-Device RAG on a Mobile NPU: System Design and Benchmark on Snapdragon X Elite}

\author{\IEEEauthorblockN{1\textsuperscript{st} Zhiyuan Cheng}
\IEEEauthorblockA{\textit{School of Engineering} \\
\textit{Stanford University}\\
Stanford, CA, USA \\
zhycheng@stanford.edu}
\and
\IEEEauthorblockN{2\textsuperscript{nd} Longying Lai}
\IEEEauthorblockA{\textit{Simon Business School} \\
\textit{University of Rochester}\\
Rochester, NY, USA \\
lai.longying@urgrad.rochester.edu}}

\maketitle

\begin{abstract}
Retrieval-Augmented Generation (RAG) pipelines are compute-intensive, combining embedding, retrieval, reranking, and large language model (LLM) generation. Running them entirely on-device benefits privacy, latency, and offline use, but the energy cost of CPU inference is a major barrier. We present what is, to our knowledge, the first end-to-end RAG pipeline that runs all neural stages---embedding, reranking, and LLM generation---on the Qualcomm Hexagon NPU of the Snapdragon X Elite. Profiling on a Dell XPS~13 laptop, we compare NPU-accelerated RAG against CPU and OpenCL/Adreno GPU baselines on indexing and query workloads. On indexing, the NPU achieves 9.1$\times$ higher embedding throughput and 12.3$\times$ less system energy. On a 120-query Wikipedia-passage benchmark, it delivers 18.1$\times$ faster LLM prefilling, 4.0$\times$ lower end-to-end query latency, and 4.0$\times$ less system energy than the CPU baseline; the same workload on the integrated GPU is 1.7$\times$ slower than CPU and uses 6.5$\times$ more energy than the NPU. A GPT-4.1 LLM-as-judge evaluation finds NPU answer quality on par with CPU and GPU within evaluator noise (mean 9.32 vs.\ 8.95 vs.\ 9.03 on a 1--10 rubric), with 86.7\% of queries scoring identically across all three backends. On the Snapdragon X Elite / Hexagon class of laptop SoC, the NPU thus enables practical, energy-efficient on-device RAG without quality regression---a sustainable path toward green edge intelligence that we expect to generalize to comparable mobile NPUs (Apple Neural Engine, Intel NPU, MediaTek APU) as their software stacks mature.
\end{abstract}


\section{Introduction}

The proliferation of large language models (LLMs) has driven demand for on-device AI systems that operate without cloud connectivity, preserving user privacy, reducing latency, and enabling offline functionality~\cite{xu2024ondevice}. Retrieval-Augmented Generation (RAG)~\cite{lewis2020rag} has emerged as the dominant paradigm for grounding LLM outputs in external knowledge, mitigating hallucination while adapting to domain-specific corpora. However, a complete RAG pipeline involves multiple compute-intensive neural inference stages---embedding generation, semantic retrieval, cross-encoder reranking, and autoregressive LLM generation---making on-device deployment challenging under the power and thermal constraints of mobile and laptop platforms.

Modern systems-on-chip (SoCs) increasingly integrate dedicated Neural Processing Units (NPUs) alongside CPUs and GPUs. Qualcomm's Snapdragon X Elite, for example, features the Hexagon NPU with up to 45~TOPS of INT8 throughput, designed specifically for sustained neural inference at low power~\cite{qualcomm2024snapdragon}. While prior work has demonstrated NPU-accelerated inference for individual models---particularly LLM prefilling~\cite{xu2025llmnpu}---we are not aware of any prior study that has implemented and benchmarked a \emph{complete} RAG pipeline on a mobile NPU.

This gap is significant for two reasons. First, a RAG pipeline loads multiple models simultaneously (embedding, reranker, LLM), creating memory allocation and scheduling challenges unique to the NPU's static computation graph architecture. Second, the energy savings from NPU offloading compound across pipeline stages, but the magnitude of this benefit has not been quantified for multi-model workloads. Understanding these trade-offs is essential for designing sustainable on-device AI systems~\cite{cheng2026toward}.

We note that on the Snapdragon X Elite, the integrated GPU (Adreno X1-85) is not a viable acceleration target for this workload: although OpenCL offload is functional, end-to-end query processing is 1.7$\times$ slower than the CPU baseline and consumes 1.6$\times$ more system energy (Section~\ref{sec:results}). This is a hardware ceiling---the integrated GPU is small relative to the X-Elite's 12 high-performance cores---not an immature software stack. The cross-encoder reranker additionally cannot run on the OpenCL backend due to a batched-scoring defect, so it stays on the NPU in all configurations. The NPU is therefore the only practical on-chip accelerator for this class of workload.

In this paper, we make the following contributions:

\begin{enumerate}
    \item \textbf{A complete NPU-accelerated RAG pipeline.} We implement what is, to our knowledge, the first end-to-end RAG system in which all three neural inference components---embedding generation (EmbeddingGemma 300M~\cite{lee2025embeddinggemma}), cross-encoder reranking (Jina Reranker v2~\cite{sturua2024jina}), and LLM generation (Qwen3-4B-Instruct~\cite{yang2025qwen3})---run on the Qualcomm Hexagon NPU through the Qualcomm AI Runtime (QAIRT/QNN) SDK~\cite{qairt}. We document critical engineering challenges including model loading order constraints, context length limitations imposed by static computation graphs, and dependency issues in the Windows ARM64 environment.

    \item \textbf{Comprehensive performance and energy benchmark.} We profile both the indexing pipeline (document parsing, embedding, FAISS index construction) and the query pipeline (hybrid retrieval, reranking, LLM generation) on identical hardware, comparing NPU against CPU baselines. We instrument power draw via HWiNFO64 shared-memory telemetry at 500\,ms granularity, measuring per-component power and total system energy.

    \item \textbf{Quantitative sustainability analysis.} We show that NPU acceleration reduces total system energy by 12.3$\times$ for indexing and 19.2$\times$ for query processing, with energy savings that are \emph{super-linear} relative to speedup because the NPU also lowers average system power draw. We contextualize these savings in terms of daily and annual energy budgets for edge RAG deployments, connecting to the Green AI agenda~\cite{schwartz2020green}.
\end{enumerate}

\section{Related Work}

\textbf{On-device LLM inference.}
The deployment of LLMs on edge devices has been enabled by quantization, pruning, and efficient runtime systems. llama.cpp~\cite{llamacpp} provides cross-platform CPU/GPU inference for GGUF-quantized models and is widely adopted for desktop and mobile deployment. MLC-LLM~\cite{mlcllm} leverages machine learning compilation to generate optimized kernels for diverse hardware backends including Vulkan and Metal. ExecuTorch~\cite{executorch} targets mobile deployment within the PyTorch ecosystem. The llm.npu system~\cite{xu2025llmnpu} is the closest predecessor to our work: it demonstrates NPU-accelerated LLM \emph{prefilling} on mobile SoCs, achieving 22.4$\times$ speedup and 30.7$\times$ energy savings over baselines by employing prompt-level chunking, tensor-level outlier extraction, and block-level heterogeneous scheduling. However, llm.npu focuses exclusively on single-model LLM inference and does not address the multi-model RAG setting. Xu et al.~\cite{xu2024ondevice} provide a comprehensive survey of on-device language model techniques, covering efficient architectures, compression, and hardware acceleration, but do not benchmark NPU-based RAG pipelines. Orthogonal to hardware offload, algorithmic efficiency techniques reduce the compute a model needs and are complementary to the NPU acceleration we study: reasoning pruning with skill-aware step decomposition~\cite{jiang2026drp}, multi-teacher distillation~\cite{zhang2026optimalteacher}, and pattern-aware tool-integrated reasoning~\cite{xu2026toolsnotjust}. Small language models can match or outperform larger ones on targeted tasks~\cite{cao2026taskspecific}, even as autonomous agents for efficient knowledge mining~\cite{zhang2026slmagents}, and data-efficiency and compression optimization further shrink deployment cost~\cite{li2024sec}; efficient sequence architectures and modular cross-domain adaptation reduce it further~\cite{li2026fast,li2025catch}; performance--efficiency trade-offs between traditional machine learning and LLMs have also been characterized directly~\cite{zhang2026performance}.

\textbf{RAG systems.}
Since the introduction of RAG by Lewis et al.~\cite{lewis2020rag}, numerous frameworks have emerged for building retrieval-augmented pipelines. LangChain~\cite{langchain} and LlamaIndex~\cite{llamaindex} provide modular abstractions for document ingestion, retrieval, and generation, but are designed primarily for server and cloud environments. Gao et al.~\cite{gao2024rag} survey the RAG landscape and identify hybrid retrieval, reranking, and iterative refinement as key design patterns. Recent work has sharpened RAG's reliability and retrieval quality: studies on knowledge-conflict and context compliance diagnose when retrieval is wrong~\cite{chen2026ragknow}, transparent knowledge-conflict handling makes such failures interpretable~\cite{ye2026seeing}, evidence-force calibration improves citation faithfulness in cited RAG~\cite{qian2026evidenceforce}, and retrieval-augmented architectures have been studied specifically under tight on-device compute budgets~\cite{liu2026architecture}. Complementary efforts mine and curate retrievable knowledge through web-scale pipelines~\cite{qi2026coverageaware} and probe how prompt design shapes LLM reasoning over retrieved evidence~\cite{qi2026correctivehints}. Retrieval itself continues to advance beyond text into composed image settings~\cite{li2025encoder,li2025finecir}, and RAG and efficient-inference techniques are increasingly applied in specialized domains such as financial decision-making~\cite{chen2026optionhedging}, multi-LLM financial sentiment forecasting~\cite{zhang2026finsentllm}, interpretable decision-support analytics~\cite{qin2025counterfactual}, and clinical and mental-health applications~\cite{yang2026llmmentalhealth}. In the financial-document setting we also evaluate, hybrid retrieval with cross-encoder reranking improves answer correctness on 10-K filings~\cite{cheng2026enhancingfinancialreportquestionanswering}, and document-routed retrieval mitigates cross-document chunk confusion in homogeneous corpora~\cite{cheng2026resolvingrobustnessprecisiontradeofffinancial}. To our knowledge, no prior work has implemented or benchmarked a complete RAG pipeline targeting on-device NPU execution.

\textbf{NPU computing and hardware acceleration.}
Neural Processing Units are increasingly integrated into consumer SoCs. Qualcomm's Hexagon NPU~\cite{qualcomm2024snapdragon} powers on-device inference on Snapdragon platforms, while Apple's Neural Engine and Intel's NPU serve similar roles on their respective architectures. Qualcomm's AI Runtime (QAIRT/QNN) SDK~\cite{qairt} exposes the Hexagon NPU to applications, compiling models into static computation graphs ahead of time; for CPU and GPU execution, llama.cpp~\cite{llamacpp} is the de facto runtime for GGUF-quantized models. Co-designing model architectures for NPU constraints, such as integer-only vision-language inference, can further narrow the hardware--model gap~\cite{chen2025autoneuralcodesigningvisionlanguagemodels}. Despite growing hardware capability, systematic benchmarks of multi-model NPU workloads remain scarce.

\textbf{Energy-efficient AI.}
Schwartz et al.~\cite{schwartz2020green} introduced the Green AI framework, advocating for energy efficiency as a first-class evaluation criterion alongside accuracy. Subsequent work has measured the carbon footprint of training large models~\cite{strubell2019energy, patterson2021carbon} and proposed carbon-aware scheduling for cloud inference. However, energy measurement for \emph{on-device} NPU inference remains underexplored, particularly for multi-stage pipelines. Our work addresses this gap by providing fine-grained power telemetry for a complete RAG workflow on a mobile NPU.

\section{System Design}

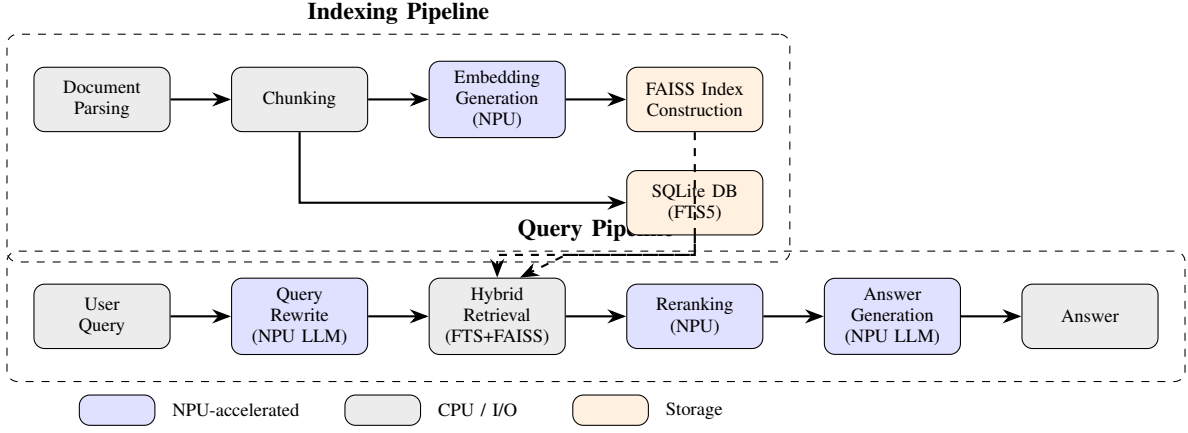
\begin{figure*}[t]
\centering
\begin{tikzpicture}[
    node distance=0.6cm and 0.8cm,
    block/.style={rectangle, draw, rounded corners, minimum height=0.85cm, minimum width=1.8cm, align=center, font=\scriptsize},
    npublock/.style={block, fill=blue!12},
    cpublock/.style={block, fill=gray!15},
    storeblock/.style={block, fill=orange!12},
    arr/.style={-{Stealth[length=2.5mm]}, thick},
    lbl/.style={font=\scriptsize\itshape, text=black!70},
]

\node[cpublock] (parse) {Document\\Parsing};
\node[cpublock, right=of parse] (chunk) {Chunking};
\node[npublock, right=of chunk] (embed) {Embedding\\Generation\\(NPU)};
\node[storeblock, right=of embed] (faiss) {FAISS Index\\Construction};
\node[storeblock, below=0.5cm of faiss] (db) {SQLite DB\\(FTS5)};

\draw[arr] (parse) -- (chunk);
\draw[arr] (chunk) -- (embed);
\draw[arr] (embed) -- (faiss);
\draw[arr] (chunk) |- (db);

\begin{scope}[on background layer]
    \node[fit=(parse)(chunk)(embed)(faiss)(db), draw, dashed, rounded corners, inner sep=0.35cm, label={[font=\small\bfseries]above:Indexing Pipeline}] (idxbox) {};
\end{scope}

\node[cpublock, below=2.0cm of parse] (query) {User\\Query};
\node[npublock, right=of query] (rewrite) {Query\\Rewrite\\(NPU LLM)};
\node[cpublock, right=of rewrite] (hybrid) {Hybrid\\Retrieval\\(FTS+FAISS)};
\node[npublock, right=of hybrid] (rerank) {Reranking\\(NPU)};
\node[npublock, right=of rerank] (llm) {Answer\\Generation\\(NPU LLM)};
\node[cpublock, right=of llm] (answer) {Answer};

\draw[arr] (query) -- (rewrite);
\draw[arr] (rewrite) -- (hybrid);
\draw[arr] (hybrid) -- (rerank);
\draw[arr] (rerank) -- (llm);
\draw[arr] (llm) -- (answer);

\draw[arr, dashed] (db) |- ([yshift=0.3cm]hybrid.north) -- (hybrid.north);
\draw[arr, dashed] (faiss) |- ([yshift=0.3cm]hybrid.north east) -- ([xshift=0.3cm]hybrid.north);

\begin{scope}[on background layer]
    \node[fit=(query)(rewrite)(hybrid)(rerank)(llm)(answer), draw, dashed, rounded corners, inner sep=0.35cm, label={[font=\small\bfseries]above:Query Pipeline}] (qrybox) {};
\end{scope}

\node[npublock, minimum width=1cm, minimum height=0.4cm, below=0.6cm of query, xshift=0.2cm] (leg1) {};
\node[right=0.1cm of leg1, font=\scriptsize] {NPU-accelerated};
\node[cpublock, minimum width=1cm, minimum height=0.4cm, right=2.5cm of leg1] (leg2) {};
\node[right=0.1cm of leg2, font=\scriptsize] {CPU / I/O};
\node[storeblock, minimum width=1cm, minimum height=0.4cm, right=2.0cm of leg2] (leg3) {};
\node[right=0.1cm of leg3, font=\scriptsize] {Storage};

\end{tikzpicture}
\caption{Architecture of the NPU-accelerated RAG system. Blue-shaded stages execute neural inference on the Hexagon NPU; gray stages run on the CPU; orange stages involve persistent storage. The indexing pipeline (top) processes documents into a searchable index; the query pipeline (bottom) retrieves relevant chunks and generates answers.}
\label{fig:architecture}
\end{figure*}

We design an end-to-end RAG system that executes all neural inference on the Qualcomm Hexagon NPU. The system comprises two pipelines---indexing and query---sharing a common model infrastructure. Fig.~\ref{fig:architecture} illustrates the overall architecture.

\subsection{Indexing Pipeline}

The indexing pipeline transforms a corpus of documents into a searchable vector index through three stages:

\textbf{Stage~1: Document parsing and chunking.}
Documents (PDF, text) are parsed into structured text using format-specific parsers. The extracted text is then split into chunks using recursive character-based splitting with a chunk size of 1{,}000 characters and 500-character overlap. These parameters are constrained by the NPU's limited context window (discussed in Section~\ref{sec:challenges}); a CPU-only deployment would use larger chunks (2{,}500 characters, 1{,}250 overlap). Chunks and metadata are stored in a SQLite database with FTS5 full-text search indexing.

\textbf{Stage~2: Embedding generation.}
Each chunk is encoded into a 1{,}024-dimensional vector using EmbeddingGemma 300M~\cite{lee2025embeddinggemma}, a lightweight embedding model from the Gemma~3 family optimized for on-device deployment. The model runs on the Hexagon NPU through the QAIRT/QNN runtime, processing chunks in batches of up to~32. Embeddings are L2-normalized before storage.

\textbf{Stage~3: FAISS index construction.}
Embedding vectors are added to a FAISS \texttt{IndexFlatL2} index wrapped with \texttt{IndexIDMap2} for chunk-ID-based retrieval~\cite{johnson2019faiss}. The flat index enables exact nearest-neighbor search, which is feasible given the on-device corpus scale (thousands to tens of thousands of chunks).

\subsection{Query Pipeline}

The query pipeline processes user queries through four stages:

\textbf{Query rewriting.}
The raw user query is optionally rewritten by the LLM (Qwen3-4B-Instruct~\cite{yang2025qwen3}) to produce a clarified query and extracted keywords. This step improves retrieval quality by resolving ambiguity and expanding query terms.

\textbf{Hybrid retrieval.}
The system performs two parallel retrieval passes: (1)~keyword-based retrieval via SQLite FTS5 (up to~20 results), and (2)~semantic retrieval via FAISS nearest-neighbor search using the query embedding (up to~30 results). Results are fused using Reciprocal Rank Fusion (RRF)~\cite{cormack2009rrf} with $k{=}60$.

\textbf{Reranking.}
Fused candidates (up to 30) are reranked by a cross-encoder model, Jina Reranker v2 Base Multilingual~\cite{sturua2024jina}, running on the NPU. The reranker assigns relevance scores; candidates are filtered using both a minimum score threshold (0.45) and a cliff-cutoff heuristic that detects sharp score drops between consecutive candidates. The top~7 chunks are retained for generation.

\textbf{Answer generation.}
The retained chunks are assembled into a context prompt and passed to Qwen3-4B-Instruct~\cite{yang2025qwen3} for streaming answer generation on the NPU. The model operates with a 4{,}096-token context window.

\subsection{Inference Backends}

Each neural component runs on one of two interchangeable backends, allowing controlled comparison across compute targets. For NPU execution, models are compiled ahead of time into static computation graphs and run through the Qualcomm AI Runtime (QAIRT/QNN) SDK~\cite{qairt}. For CPU and GPU execution, the same model architectures are run in GGUF-quantized format through llama.cpp~\cite{llamacpp}, with GPU offload enabled by setting \texttt{n\_gpu\_layers} to the full layer count. We wrap both backends behind a common interface so that the embedding, reranking, and LLM stages can be independently assigned to the NPU, CPU, or GPU without changing pipeline logic.

\subsection{Engineering Challenges}
\label{sec:challenges}

Deploying a multi-model RAG pipeline on the NPU revealed several non-trivial engineering challenges:

\textbf{Model loading order.}
The NPU requires models to be loaded in descending order of size: LLM (4B parameters) first, then embedding (300M), and finally reranker (278M). Violating this order causes static computation graph memory allocation failures, likely due to the NPU's contiguous memory allocation strategy. This constraint does not apply to CPU/GPU backends.

\textbf{Context length limitations.}
NPU static graphs impose a fixed maximum context length, which is more restrictive than the flexible KV-cache approach used in CPU/GPU inference. This necessitates smaller chunk sizes during indexing (1{,}000 vs.\ 2{,}500 characters) and fewer assembled chunks during generation (7 vs.\ 10), reducing the context available to the LLM but remaining sufficient for our benchmarks.

\textbf{ARM64 Windows ecosystem.}
The entire pipeline runs in a native Windows ARM64 Python environment on the Snapdragon X Elite platform. Many Python packages lack pre-built ARM64 Windows wheels, requiring manual compilation or alternative dependencies. This ecosystem immaturity adds substantial development overhead but is essential for leveraging the NPU through the QAIRT/QNN Windows ARM64 runtime.

\section{Experimental Setup}

\subsection{Hardware Platform}

All experiments are conducted on a Dell XPS~13 9345 laptop. Table~\ref{tab:hardware} summarizes the hardware specifications. To ensure measurement integrity, no other applications run during experiments, and the system operates on AC power in battery performance mode.

\begin{table}[t]
\centering
\caption{Hardware specifications of the test platform.}
\label{tab:hardware}
\begin{tabular}{@{}ll@{}}
\toprule
\textbf{Component} & \textbf{Specification} \\
\midrule
Device & Dell XPS 13 9345 \\
CPU & Snapdragon X Elite X1E80100 \\
    & (Qualcomm Oryon, 12 cores) \\
GPU & Qualcomm Adreno X1-85 \\
NPU & Hexagon NPU (45 TOPS INT8) \\
Memory & 64\,GB LPDDR5x \\
OS & Windows 11 ARM64 \\
Power supply & 100\,W AC adapter \\
Power mode & Battery performance mode \\
\bottomrule
\end{tabular}
\end{table}

\subsection{Models}

Table~\ref{tab:models} lists the models used in the RAG pipeline. For NPU execution, models are compiled ahead of time into the QAIRT/QNN static-graph format. For the CPU and GPU baselines, the same model architectures are used in GGUF-quantized format, inferred via llama.cpp.

\begin{table}[t]
\centering
\caption{Models used in the RAG pipeline.}
\label{tab:models}
\begin{tabular}{@{}llll@{}}
\toprule
\textbf{Component} & \textbf{Model} & \textbf{Params} & \textbf{Dim} \\
\midrule
Embedding & EmbeddingGemma~\cite{lee2025embeddinggemma} & 300M & 1024 \\
Reranker & Jina Reranker v2~\cite{sturua2024jina} & 278M & --- \\
LLM & Qwen3-4B-Instruct~\cite{yang2025qwen3} & 4B & --- \\
\bottomrule
\end{tabular}
\end{table}

\subsection{Dataset and Workload}

For \textbf{indexing}, we process a corpus of 10~documents (the SEC 10-K filings of A, AAL, AAPL, ABBV, ABNB, ABT, ACGL, ACN, ADBE, and ADI), producing 9{,}324~chunks comprising approximately 8.14~million characters ($\approx$2.04~million estimated tokens).

For \textbf{query processing}, we benchmark on two complementary 120-query test sets, each requiring full pipeline execution (retrieval, reranking, and generation):
\begin{itemize}
    \item \textbf{wiki\_minirag} (general-knowledge factoid QA). 3{,}187 Wikipedia passages drawn from the \texttt{rag-mini-wikipedia} dataset~\cite{ragminiwikipedia}; 120 queries deterministically sampled (every 7th row) from the dataset's 918-question test split. Wikipedia passages contain the answer to every query, allowing the benchmark to isolate backend-induced quality differences from corpus-coverage artifacts. We use this set as the primary workload for performance, energy, and answer-quality evaluation.
    \item \textbf{FinDER} (domain-specialized financial QA). 120 queries from the FinDER~\cite{linq2024finder} test split, restricted to the 10~indexed companies. Queries are heavily abbreviated (e.g., ``ADI's eng.\ talent ratio \& turnover impact on comp.\ innovation.''); we use this set as a stress test of correct refusal behavior on a tiny domain-specialist corpus.
\end{itemize}

\subsection{Profiling Methodology}

We instrument profiling at three levels:

\textbf{Stage-level timing.}
Wall-clock time for each pipeline stage is measured using Python's \texttt{time.perf\_counter()}, providing microsecond-resolution timestamps. Throughput is computed as tokens processed per second for embedding (prefilling speed) and LLM inference (prefilling and decoding speeds separately).

\textbf{Power and energy monitoring.}
A background monitoring thread samples system telemetry at 500\,ms intervals throughout each experiment. On the Windows platform, we read HWiNFO64 sensor data from shared memory (\texttt{Global\textbackslash HWiNFO\_SENS\_SM2}), capturing per-component power draw (CPU clusters 0/1/2, GPU, system total), NPU utilization percentage, and GPU utilization. System-level resource metrics (CPU utilization, process memory) are collected via \texttt{psutil}. Total energy consumption (Joules) is computed by numerical integration of instantaneous power samples over the experiment duration.

\textbf{LLM profiling.}
For query processing, we measure time-to-first-token (TTFT) to isolate the prefilling phase, and tokens-per-second during decoding. These metrics are extracted from the runtime's built-in profiling interface (QAIRT/QNN for the NPU, llama.cpp for the CPU and GPU).

\textbf{Answer-quality evaluation.}
We score every generated answer against a per-query reference answer using GPT-4.1 as an LLM-as-judge~\cite{zheng2023llmjudge,li2025judgesurvey}, with temperature~0.0 and a 1--10 rubric where $\geq$7 denotes a materially correct answer and 10 denotes a perfect match. The judge sees only the question, the system's answer, and the reference; it does not see which backend produced the answer. We report mean score, failure rate ($s{=}1$), correctness rate ($s{\geq}7$), and perfect rate ($s{=}10$) for each backend, plus paired exact Wilcoxon signed-rank tests on the per-query score differences. We use a single judge in line with established LLM-as-judge practice for systems papers; this caveat is discussed in Section~\ref{sec:limitations}.

\subsection{Baselines}

Our experiments compare three on-chip backends on the same Snapdragon X Elite SoC, using identical model architectures and identical pipeline code. The reranker stays on the Hexagon NPU in all configurations (the OpenCL backend exhibits a batched-scoring defect, and pinning the reranker isolates the LLM as the only varying component). The three backends differ only in where the LLM (and embedder) executes:
\begin{itemize}
    \item \textbf{NPU.} Embedding, reranking, and LLM generation all execute on the Hexagon NPU through the QAIRT/QNN SDK using ahead-of-time-compiled static-graph models.
    \item \textbf{CPU.} The LLM and embedder run on the X-Elite CPU through llama.cpp with \texttt{n\_gpu\_layers=0}, using GGUF Q4\_K\_M weights for the LLM and BF16 weights for the embedder. The reranker remains on the NPU.
    \item \textbf{GPU.} The LLM and embedder run on the Adreno X1-85 integrated GPU through llama.cpp with full OpenCL offload (\texttt{n\_gpu\_layers=999}), using the same GGUF/BF16 weights as the CPU baseline. The reranker remains on the NPU.
\end{itemize}
This design ensures that performance, energy, and quality differences across the three rows are attributable to the LLM compute backend rather than to model architecture, quantization, or pipeline configuration.

\section{Results and Analysis}
\label{sec:results}

\subsection{Indexing Performance}

Table~\ref{tab:indexing} presents the indexing benchmark results. The NPU achieves a \textbf{9.1$\times$} improvement in embedding throughput (3{,}325 vs.\ 367~tokens/s), which dominates the total pipeline time. Stage~1 (parsing/chunking) and Stage~3 (FAISS construction) show similar performance across backends, as expected for CPU-bound and I/O-bound operations respectively.

\begin{table}[t]
\centering
\caption{Indexing performance comparison (10 documents, 9{,}324 chunks, $\approx$2.04M tokens).}
\label{tab:indexing}
\begin{tabular}{@{}lrrr@{}}
\toprule
\textbf{Metric} & \textbf{NPU} & \textbf{CPU} & \textbf{Ratio} \\
\midrule
\multicolumn{4}{@{}l}{\textit{Stage Timing (seconds)}} \\
\quad Parse \& chunk & 97.6 & 92.7 & 1.0$\times$ \\
\quad Embedding & 612.4 & 5{,}554.5 & 9.1$\times$ \\
\quad FAISS build & 0.5 & 0.8 & 1.6$\times$ \\
\quad Total pipeline & 710.7 & 5{,}648.2 & 7.9$\times$ \\
\midrule
\multicolumn{4}{@{}l}{\textit{Embedding Throughput}} \\
\quad Tokens/s & 3{,}325 & 367 & 9.1$\times$ \\
\quad Chunks/s & 15.2 & 1.7 & 8.9$\times$ \\
\midrule
\multicolumn{4}{@{}l}{\textit{Resource Utilization}} \\
\quad Avg CPU util.\ (\%) & 12.6 & 98.5 & --- \\
\quad Avg NPU util.\ (\%) & 84.2 & 0.0 & --- \\
\midrule
\multicolumn{4}{@{}l}{\textit{Energy}} \\
\quad Avg system power (W) & 27.6 & 42.7 & 0.65$\times$ \\
\quad Total system energy (J) & 19{,}592 & 241{,}020 & \textbf{12.3$\times$} \\
\bottomrule
\end{tabular}
\end{table}

The total pipeline speedup (7.9$\times$) is lower than the embedding-stage speedup (9.1$\times$) because the parsing stage constitutes a fixed overhead that is identical across backends. With embedding dominating pipeline time (86\% for NPU, 98\% for CPU), the NPU effectively eliminates the embedding bottleneck.

\subsection{Query Performance}

Table~\ref{tab:query} summarizes query processing results on the wiki\_minirag benchmark, averaged over 120~queries, for all three on-chip backends. The NPU delivers dramatic improvements over both CPU and GPU baselines on the LLM-intensive stages.

\begin{table}[t]
\centering
\caption{Query performance on wiki\_minirag (mean over 120~queries). Ratios are taken against the CPU baseline.}
\label{tab:query}
\small
\setlength{\tabcolsep}{4pt}
\begin{tabular}{@{}lrrrrr@{}}
\toprule
\textbf{Metric} & \textbf{NPU} & \textbf{CPU} & \textbf{GPU} & \textbf{N/C} & \textbf{G/C} \\
\midrule
\multicolumn{6}{@{}l}{\textit{Latency (seconds)}} \\
\quad Retrieval time     &   3.09 &   3.09 &   3.10 & 1.0$\times$  & 1.0$\times$  \\
\quad LLM TTFT           &   1.30 &  24.77 &  42.24 & 19.1$\times$ & 0.59$\times$ \\
\quad Total query time   &   9.48 &  37.98 &  63.61 & 4.0$\times$  & 0.60$\times$ \\
\midrule
\multicolumn{6}{@{}l}{\textit{LLM Throughput (tokens/s)}} \\
\quad Prefilling speed   & 786.66 &  43.36 &  25.18 & 18.1$\times$ & 0.58$\times$ \\
\quad Decoding speed     &  14.19 &   8.17 &   4.65 & 1.74$\times$ & 0.57$\times$ \\
\midrule
\multicolumn{6}{@{}l}{\textit{Resource Utilization}} \\
\quad Avg CPU util.\ (\%)&  55.84 &  79.10 &  41.86 & --- & --- \\
\quad Avg NPU util.\ (\%)&  97.30 &   7.67 &   4.62 & --- & --- \\
\quad Avg GPU D3D (\%)   &  10.37 &   9.83 &  71.19 & --- & --- \\
\midrule
\multicolumn{6}{@{}l}{\textit{Tail Latency (P95, seconds)}} \\
\quad Total query time   &  17.90 &  69.56 & 107.40 & 3.9$\times$  & 0.65$\times$ \\
\midrule
\multicolumn{6}{@{}l}{\textit{Energy (total over 120 queries)}} \\
\quad Avg sys.\ power (W)&  33.24 &  32.94 &  32.25 & 1.01$\times$ & 0.98$\times$ \\
\quad Total sys.\ E (kJ) &  37.83 & 150.12 & 246.14 & \textbf{4.0$\times$} & 0.61$\times$ \\
\bottomrule
\end{tabular}
\end{table}

\textbf{Prefilling speed} is the standout result: the NPU achieves 786.7~tokens/s, an 18.1$\times$ improvement over the CPU's 43.4~tokens/s and a 31.2$\times$ improvement over the GPU's 25.2~tokens/s. This aligns with the NPU's strength in dense matrix-multiplication workloads that dominate the prefilling phase. \textbf{Decoding speed} shows a more modest 1.74$\times$ improvement over CPU (14.2 vs.\ 8.2~tokens/s) and 3.05$\times$ over GPU, consistent with the memory-bandwidth-bound, inherently sequential nature of autoregressive token generation.

\textbf{The Adreno X1-85 GPU is genuinely active but slower than the X-Elite CPU} on this 4B Q4\_K\_M model. The GPU configuration sustains 71.2\% GPU D3D utilization throughout the run (vs.\ $\approx$10\% in the NPU and CPU configurations), confirming OpenCL offload is functional. However, the integrated GPU is small relative to the X-Elite's 12 high-performance cores, and decoding is 1.7$\times$ slower than CPU. This is a hardware-ceiling result, not a runtime defect.

\textbf{Retrieval time} is nearly identical across backends ($\approx$3.1\,s), as expected: this stage involves FAISS vector search and SQLite FTS5 queries, which are CPU-bound and do not benefit from NPU or GPU acceleration.

\textbf{Tail latency} (P95) reflects the same ranking as the means: 17.9\,s NPU vs.\ 69.6\,s CPU vs.\ 107.4\,s GPU. The NPU provides consistent, predictable performance suitable for interactive use.

\subsection{Energy Analysis}

\begin{table}[t]
\centering
\caption{Per-component energy breakdown over the 120-query wiki\_minirag batch (HWiNFO64 sensors, 500\,ms sampling).}
\label{tab:power}
\small
\setlength{\tabcolsep}{3pt}
\begin{tabular}{@{}lrrrrrr@{}}
\toprule
 & \multicolumn{2}{c}{\textbf{NPU}} & \multicolumn{2}{c}{\textbf{CPU}} & \multicolumn{2}{c}{\textbf{GPU}} \\
\cmidrule(lr){2-3} \cmidrule(lr){4-5} \cmidrule(lr){6-7}
\textbf{Component} & \textbf{W} & \textbf{kJ} & \textbf{W} & \textbf{kJ} & \textbf{W} & \textbf{kJ} \\
\midrule
CPU Cluster 0 & 4.11  &  4.67 & 5.33  & 24.30 & 3.33  & 25.44 \\
CPU Cluster 1 & 4.86  &  5.53 & 5.48  & 24.99 & 3.89  & 29.71 \\
CPU Cluster 2 & 4.63  &  5.27 & 6.46  & 29.46 & 4.11  & 31.40 \\
GPU           & 0.08  &  0.10 & 0.07  &  0.33 & 2.15  & 16.40 \\
System total  & 33.24 & 37.83 & 32.94 & 150.12 & 32.25 & 246.14 \\
\bottomrule
\end{tabular}
\end{table}

Table~\ref{tab:power} shows the per-component energy breakdown during query processing. \textbf{Average system power is essentially identical across all three backends} (33.24, 32.94, 32.25\,W); the platform's idle baseline dominates the system rail. The energy ranking is therefore set almost entirely by wall-clock time: NPU completes the 120-query batch in 19\,min, CPU in 76\,min, GPU in 127\,min.

For indexing, the original 7.9$\times$ NPU/CPU speedup yields a 12.3$\times$ system-energy reduction through the more pronounced super-linear dual effect we documented in earlier experiments: with the NPU handling embedding, CPU clusters operate at near-idle power, reducing average system power by 35\% (27.6\,W vs.\ 42.7\,W). Indexing is a substantially more compute-dense workload than per-query LLM generation in the on-device setting, which is why the system-power gap is wider for indexing.

For query processing, the LLM-decoding phase is memory-bandwidth-bound; the system power difference between backends collapses to noise, so the energy ratio tracks the latency ratio (4.0$\times$ NPU/CPU, 0.61$\times$ GPU/CPU, 6.5$\times$ NPU/GPU). The OpenCL run consumes 16.4\,kJ in the GPU domain alone---165$\times$ the NPU run's GPU energy---confirming Adreno is genuinely active, but it does not translate into wall-clock savings.

\begin{figure}[t]
\centering
\begin{tikzpicture}
\begin{scope}[xscale=0.62]
    \draw[fill=blue!25] (0,0)   rectangle (1.4, 1.96) node[midway, font=\scriptsize] {19.6};
    \draw[fill=red!25]  (1.8,0) rectangle (3.2, 4.5)  node[midway, font=\scriptsize] {241.0};

    \draw[fill=blue!25]   (5.0,0) rectangle (6.4, 0.71) node[midway, font=\tiny] {37.8};
    \draw[fill=red!25]    (6.8,0) rectangle (8.2, 2.80) node[midway, font=\scriptsize] {150.1};
    \draw[fill=orange!35] (8.6,0) rectangle (10.0, 4.59) node[midway, font=\scriptsize] {246.1};

    \node[font=\scriptsize] at (0.7, -0.35) {NPU};
    \node[font=\scriptsize] at (2.5, -0.35) {CPU};
    \node[font=\scriptsize] at (5.7, -0.35) {NPU};
    \node[font=\scriptsize] at (7.5, -0.35) {CPU};
    \node[font=\scriptsize] at (9.3, -0.35) {GPU};

    \node[font=\small\bfseries] at (1.6, -0.95) {Indexing};
    \node[font=\small\bfseries] at (7.5, -0.95) {Query (120)};

    \draw[<->, thick] (3.4, 1.96) -- (3.4, 4.5) node[midway, right, font=\tiny] {12.3$\times$};
    \draw[<->, thick] (10.2, 0.71) -- (10.2, 4.59) node[midway, right, font=\tiny] {6.5$\times$};
\end{scope}
\end{tikzpicture}
\caption{Total system energy in kJ (HWiNFO64). Indexing: NPU vs.\ CPU. Query: NPU vs.\ CPU vs.\ GPU/OpenCL on the wiki\_minirag 120-query benchmark.}
\label{fig:energy}
\end{figure}
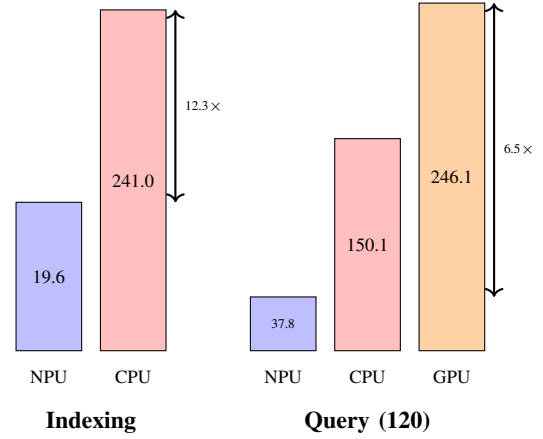

Fig.~\ref{fig:energy} visualizes total system energy across indexing (NPU vs.\ CPU) and query processing (all three backends).

\subsection{RAG Answer Quality}
\label{sec:quality}

The NPU runs a deliberately constrained configuration: 1{,}000-character chunks (vs.\ 2{,}500 on CPU/GPU), top-7 retrieved chunks (vs.\ top-10), and a 4{,}096-token context window. A natural concern is whether this constraint degrades answer quality. Table~\ref{tab:quality_wiki} reports a GPT-4.1 LLM-as-judge evaluation on the wiki\_minirag test set across all three backends.

\begin{table}[t]
\centering
\caption{Answer quality on wiki\_minirag (GPT-4.1 judge, 1--10 rubric, 120 queries per backend).}
\label{tab:quality_wiki}
\small
\setlength{\tabcolsep}{4pt}
\begin{tabular}{@{}lrrrr@{}}
\toprule
\textbf{Backend} & \textbf{Avg} & \textbf{Fail ($s{=}1$)} & \textbf{Correct ($s{\geq}7$)} & \textbf{Perfect ($s{=}10$)} \\
\midrule
CPU         & 8.95 & 5.0\% & 87.5\% & 84.2\% \\
GPU/OpenCL  & 9.03 & 6.7\% & 88.3\% & 85.0\% \\
NPU         & 9.32 & 5.0\% & 92.5\% & 88.3\% \\
\bottomrule
\end{tabular}
\end{table}

\textbf{Quality is on par across all three backends.} 104 of 120 queries (86.7\%) receive identical scores from all three backends. Pairwise paired exact Wilcoxon signed-rank tests on the $\leq$15 discordant pairs yield $p{=}0.039$ for NPU\,$-$\,CPU (mean diff $+0.367$), $p{=}0.121$ for NPU\,$-$\,GPU ($+0.283$), and $p{=}0.313$ for CPU\,$-$\,GPU ($-0.083$). A manual spot-check of the largest disagreements shows the residual gap is dominated by judge sensitivity rather than substantive backend differences---e.g., on the query \emph{``What is the biggest city in Finland?''} the CPU and GPU emit nearly identical text, yet are scored 2 and 8 respectively, a 6-point swing on essentially the same answer; on \emph{``Who did Ford nominate for Vice President?''} all three backends correctly answer ``Nelson Rockefeller'' (the figure named in the retrieved passage), but the dataset's reference is ``Bob Dole'' (a different event), so two of the three direct answers receive a~1 from the judge. We therefore report this as \emph{answer-quality parity within evaluator noise}: the NPU's constrained configuration does not materially degrade quality, even as it delivers a 4.0$\times$ system-energy reduction.

\textbf{Refusal behavior on a domain-specialist corpus.} The FinDER queries are heavily abbreviated and target a tiny 10-document corpus, so we use this dataset to evaluate \emph{correct refusal} rather than answer quality. Across the 120-query test set, the NPU returns substantive answers for 9.2\%, refusing the rest with messages such as ``the provided context does not contain sufficient information''; the CPU returns substantive answers for 44.2\%. Among the substantive subset, mean GPT-4.1 scores are 6.45 (NPU) and 7.47 (CPU). Both backends therefore refuse rather than fabricate when retrieval evidence is weak---the desired conservative behavior for an on-device assistant. The substantive-rate gap reflects the CPU's larger context window (top-10 chunks of 2{,}500 characters) admitting more borderline-relevant passages, not a quality difference; in line with the wiki\_minirag finding, the answers that are produced are of comparable quality.

\section{Discussion}

\subsection{Sustainability Implications}

The energy savings demonstrated by NPU-accelerated RAG have meaningful implications for sustainable AI deployment at the edge. Per-query system energy on the wiki\_minirag benchmark is 315\,J for the NPU, 1{,}251\,J for the CPU, and 2{,}051\,J for the GPU. Consider a scenario in which an on-device RAG system processes 1{,}000 queries per day:

\begin{itemize}
    \item \textbf{NPU:}        $1{,}000 \times 315\,\text{J}    = 315\,\text{kJ/day}    \approx 87.5\,\text{Wh/day}$
    \item \textbf{CPU baseline:} $1{,}000 \times 1{,}251\,\text{J} = 1{,}251\,\text{kJ/day} \approx 347.5\,\text{Wh/day}$
    \item \textbf{GPU/OpenCL:}  $1{,}000 \times 2{,}051\,\text{J} = 2{,}051\,\text{kJ/day} \approx 569.7\,\text{Wh/day}$
\end{itemize}

Relative to the CPU baseline, the NPU saves approximately 260\,Wh per day, or 94.9\,kWh per year per device. For fleet deployments across thousands of edge devices, these savings scale proportionally, contributing to measurable reductions in operational carbon footprint. The integrated GPU is, in this regime, the \emph{worst} of the three backends---an important data point for practitioners who might otherwise default to GPU offload because it is the conventional choice on x86 platforms.

Beyond energy, when LLM inference moves to the NPU the CPU clusters drop to near-idle power (4.1--4.6\,W average vs.\ 5.3--6.5\,W under CPU inference), freeing CPU capacity for other on-device tasks. Lower thermal dissipation also reduces fan noise and extends device longevity, both of which matter for battery-powered laptop and tablet form factors.

\subsection{Limitations and Future Work}
\label{sec:limitations}

Our study has several limitations that suggest directions for future research:

\textbf{Static computation graphs.}
The NPU's requirement for statically compiled models imposes fixed context length limits, preventing dynamic context expansion. Future NPU runtimes with dynamic shape support would alleviate the chunk size and context window constraints documented in Section~\ref{sec:challenges}.

\textbf{Model loading constraints.}
The observed largest-first model loading order requirement reflects current NPU memory management limitations. As NPU software stacks mature, we expect more flexible memory allocation strategies to emerge.

\textbf{Limited model ecosystem.}
Not all model architectures have NPU-compiled variants available. Expanding the set of NPU-compatible embedding, reranking, and LLM models would broaden the applicability of NPU-accelerated RAG.

\textbf{Single-device evaluation.}
Our benchmarks are conducted on a single hardware platform (Snapdragon X Elite). Evaluating across multiple NPU architectures (Apple Neural Engine, Intel NPU, MediaTek APU) and form factors (smartphones, tablets) would strengthen generalizability. We expect the qualitative finding---that an NPU accelerates the prefilling-dominated phase of RAG enough to make end-to-end on-device deployment energy-competitive---to carry across architectures, but the specific speedup and energy ratios are platform-dependent.

\textbf{Single LLM-as-judge evaluator.}
Our quality evaluation in Section~\ref{sec:quality} uses one judge (GPT-4.1, temperature~0.0) and no human annotation. A multi-evaluator cross-check or human-graded subset would harden the parity claim. The spot-check in Section~\ref{sec:quality} shows that the residual NPU--CPU gap is dominated by judge sensitivity rather than substantive backend differences, which is consistent with the conclusion (parity within evaluator noise) but does not by itself rule out judge bias~\cite{li2025judgesurvey}. More broadly, ensuring LLM and vision-language outputs are robust and safe remains an active concern~\cite{lin2026reflectguard,wang2026visualleakbench}, including the compositional risks that arise when agent skills are combined~\cite{wang2026safeskills}.

\section{Conclusion}

We have presented what is, to our knowledge, the first end-to-end RAG pipeline that executes all neural inference stages on a mobile NPU, demonstrating the viability of NPU-accelerated on-device retrieval-augmented generation. Our system integrates embedding generation, cross-encoder reranking, and LLM-based answer generation on the Qualcomm Hexagon NPU of the Snapdragon X Elite through the Qualcomm AI Runtime (QAIRT/QNN) SDK, with llama.cpp providing the CPU and GPU baselines.

Comprehensive benchmarks on a Dell XPS~13 laptop reveal that NPU acceleration delivers 9.1$\times$ higher embedding throughput and 18.1$\times$ faster LLM prefilling compared to CPU inference, while reducing total system energy by 12.3$\times$ for indexing and 4.0$\times$ for query processing on a 120-query Wikipedia-passage benchmark. A direct comparison against the same Snapdragon X Elite's integrated GPU shows that OpenCL offload to the Adreno X1-85, while functional, is 1.7$\times$ slower than the CPU on this workload and consumes 6.5$\times$ more system energy than the NPU---an important data point for practitioners who might default to GPU offload as the conventional accelerator choice. A GPT-4.1 LLM-as-judge evaluation confirms that the NPU's constrained configuration produces answers of equivalent quality to the CPU and GPU within evaluator noise, indicating that the energy savings come essentially for free on quality. These results position the Hexagon-class mobile NPU as a compelling platform for energy-efficient on-device AI on the Snapdragon X Elite, with a system design that we expect to carry over to comparable mobile NPU architectures (Apple Neural Engine, Intel NPU, MediaTek APU) as their hardware capabilities, software toolchains, and model ecosystems continue to mature.

Our source code will be made publicly available at \url{https://github.com/zhycheng614/NPU-RAG-public} upon publication of this work.

\bibliographystyle{IEEEtran}
\bibliography{references}

\begin{thebibliography}{10}
\providecommand{\url}[1]{#1}
\csname url@samestyle\endcsname
\providecommand{\newblock}{\relax}
\providecommand{\bibinfo}[2]{#2}
\providecommand{\BIBentrySTDinterwordspacing}{\spaceskip=0pt\relax}
\providecommand{\BIBentryALTinterwordstretchfactor}{4}
\providecommand{\BIBentryALTinterwordspacing}{\spaceskip=\fontdimen2\font plus
\BIBentryALTinterwordstretchfactor\fontdimen3\font minus
  \fontdimen4\font\relax}
\providecommand{\BIBforeignlanguage}[2]{{%
\expandafter\ifx\csname l@#1\endcsname\relax
\typeout{** WARNING: IEEEtran.bst: No hyphenation pattern has been}%
\typeout{** loaded for the language `#1'. Using the pattern for}%
\typeout{** the default language instead.}%
\else
\language=\csname l@#1\endcsname
\fi
#2}}
\providecommand{\BIBdecl}{\relax}
\BIBdecl

\bibitem{xu2024ondevice}
J.~Xu \emph{et~al.}, ``On-device language models: A comprehensive review,''
  \emph{arXiv preprint arXiv:2409.00088}, 2024.

\bibitem{lewis2020rag}
P.~Lewis, E.~Perez, A.~Piktus, F.~Petroni, V.~Karpukhin, N.~Goyal,
  H.~K{\"u}ttler, M.~Lewis, W.-t. Yih, T.~Rockt{\"a}schel, S.~Riedel, and
  D.~Kiela, ``Retrieval-augmented generation for knowledge-intensive {NLP}
  tasks,'' in \emph{Advances in Neural Information Processing Systems},
  vol.~33, 2020, pp. 9459--9474.

\bibitem{qualcomm2024snapdragon}
{Qualcomm Technologies, Inc.}, ``Snapdragon {X} elite platform,''
  \url{https://www.qualcomm.com/products/mobile/snapdragon/pcs-and-tablets/snapdragon-x-elite},
  2024.

\bibitem{xu2025llmnpu}
D.~Xu, H.~Zhang, L.~Yang, R.~Liu, G.~Huang, M.~Xu, and X.~Liu, ``Fast on-device
  {LLM} inference with {NPUs},'' in \emph{Proceedings of the 30th ACM
  International Conference on Architectural Support for Programming Languages
  and Operating Systems (ASPLOS '25), Volume 1}.\hskip 1em plus 0.5em minus
  0.4em\relax ACM, 2025, pp. 445--462.

\bibitem{cheng2026toward}
Z.~Cheng, L.~Lai, Y.~Liu, and Y.~Sun, ``Toward sustainable on-device
  intelligence: A survey on energy-efficient {RAG} systems with small language
  models,'' \emph{Available at SSRN 6698538}, 2026.

\bibitem{lee2025embeddinggemma}
J.~Lee \emph{et~al.}, ``{EmbeddingGemma}: Powerful and lightweight text
  representations,'' \emph{arXiv preprint arXiv:2509.20354}, 2025.

\bibitem{sturua2024jina}
N.~Sturua \emph{et~al.}, ``jina-reranker-v2-base-multilingual,''
  \url{https://huggingface.co/jinaai/jina-reranker-v2-base-multilingual}, 2024,
  jina AI.

\bibitem{yang2025qwen3}
A.~Yang \emph{et~al.}, ``Qwen3 technical report,'' \emph{arXiv preprint
  arXiv:2505.09388}, 2025.

\bibitem{qairt}
{Qualcomm Technologies, Inc.}, ``Qualcomm {AI} runtime ({QAIRT}) {SDK},''
  \url{https://www.qualcomm.com/developer/software/qualcomm-ai-runtime-sdk},
  2024.

\bibitem{schwartz2020green}
R.~Schwartz, J.~Dodge, N.~A. Smith, and O.~Etzioni, ``Green {AI},''
  \emph{Communications of the ACM}, vol.~63, no.~12, pp. 54--63, 2020.

\bibitem{llamacpp}
G.~Gerganov \emph{et~al.}, ``llama.cpp: Inference of {Meta's LLaMA} model in
  pure {C/C++},'' \url{https://github.com/ggerganov/llama.cpp}, 2024.

\bibitem{mlcllm}
{MLC AI}, ``{MLC-LLM}: Machine learning compilation for large language
  models,'' \url{https://llm.mlc.ai/}, 2024.

\bibitem{executorch}
{Meta Platforms, Inc.}, ``{ExecuTorch}: End-to-end solution for enabling
  on-device inference,'' \url{https://github.com/pytorch/executorch}, 2024.

\bibitem{jiang2026drp}
\BIBentryALTinterwordspacing
Y.~Jiang, D.~Li, and F.~Ferraro, ``{DRP}: Distilled reasoning pruning with
  skill-aware step decomposition for efficient large reasoning models,'' 2026.
  [Online]. Available: \url{https://arxiv.org/abs/2505.13975}
\BIBentrySTDinterwordspacing

\bibitem{zhang2026optimalteacher}
\BIBentryALTinterwordspacing
H.~Zhang, S.~Yang, X.~Liang, C.~Shang, Y.~Jiang, C.~Tao, J.~Xiong, H.~K.-H. So,
  R.~Xie, A.~X. Chang, and N.~Wong, ``Find your optimal teacher: Personalized
  data synthesis via router-guided multi-teacher distillation,'' 2026.
  [Online]. Available: \url{https://arxiv.org/abs/2510.10925}
\BIBentrySTDinterwordspacing

\bibitem{xu2026toolsnotjust}
\BIBentryALTinterwordspacing
N.~Xu, Y.~Jiang, S.~R. Dipta, and H.~Zhang, ``Learning how to use tools, not
  just when: Pattern-aware tool-integrated reasoning,'' 2026. [Online].
  Available: \url{https://arxiv.org/abs/2509.23292}
\BIBentrySTDinterwordspacing

\bibitem{cao2026taskspecific}
\BIBentryALTinterwordspacing
J.~Cao, Y.~Ma, X.~Li, Q.~Ren, and X.~Chen, ``Task-specific efficiency analysis:
  When small language models outperform large language models,'' 2026.
  [Online]. Available: \url{https://arxiv.org/abs/2603.21389}
\BIBentrySTDinterwordspacing

\bibitem{zhang2026slmagents}
\BIBentryALTinterwordspacing
S.~Zhang, S.~Lin, X.~Wei, Y.~Chen, P.~Qian, S.~Wang, and H.~Xu, ``Small
  language model agents enable efficient and high-quality knowledge mining,''
  2026. [Online]. Available: \url{https://arxiv.org/abs/2510.01427}
\BIBentrySTDinterwordspacing

\bibitem{li2024sec}
X.~Li, Y.~Ma, Y.~Huang, X.~Wang, Y.~Lin, and C.~Zhang, ``Synergized data
  efficiency and compression ({SEC}) optimization for large language models,''
  in \emph{2024 4th International Conference on Electronic Information
  Engineering and Computer Science (EIECS)}.\hskip 1em plus 0.5em minus
  0.4em\relax IEEE, 2024, pp. 586--591.

\bibitem{li2026fast}
\BIBentryALTinterwordspacing
X.~Li, J.~Cao, M.~Wang, Y.~Wu, L.~Yan, Y.~Zhou, Z.~Sha, and Y.~Ma, ``{FAST}: A
  synergistic framework of attention and state-space models for spatiotemporal
  traffic prediction,'' 2026. [Online]. Available:
  \url{https://arxiv.org/abs/2604.13453}
\BIBentrySTDinterwordspacing

\bibitem{li2025catch}
X.~Li, Y.~Lu, J.~Cao, Y.~Ma, Z.~Li, and Y.~Zhou, ``{CATCH}: A modular
  cross-domain adaptive template with hook,'' in \emph{International Symposium
  on Visual Computing}.\hskip 1em plus 0.5em minus 0.4em\relax Springer, 2025,
  pp. 41--52.

\bibitem{zhang2026performance}
Y.~Zhang, Z.~Xiang, and H.~Xu, ``Performance-efficiency trade-offs in human
  preference prediction: A comparative study of traditional machine learning
  and large language models,'' in \emph{Proceedings of the 31st IEEE Symposium
  on Computers and Communications (ISCC)}.\hskip 1em plus 0.5em minus
  0.4em\relax IEEE, 2026.

\bibitem{langchain}
{LangChain, Inc.}, ``{LangChain}: Building applications with {LLMs} through
  composability,'' \url{https://github.com/langchain-ai/langchain}, 2024.

\bibitem{llamaindex}
{LlamaIndex}, ``{LlamaIndex}: A data framework for {LLM} applications,''
  \url{https://github.com/run-llama/llama_index}, 2024.

\bibitem{gao2024rag}
Y.~Gao, Y.~Xiong, X.~Gao, K.~Jia, J.~Pan, Y.~Bi, Y.~Dai, J.~Sun, and H.~Wang,
  ``Retrieval-augmented generation for large language models: A survey,''
  \emph{arXiv preprint arXiv:2312.10997}, 2024.

\bibitem{chen2026ragknow}
\BIBentryALTinterwordspacing
Y.~Chen, P.~Qian, S.~Wang, S.~Zhang, H.~Xu, S.~Lin, and X.~Wei, ``Does {RAG}
  know when retrieval is wrong? diagnosing context compliance under knowledge
  conflict,'' 2026. [Online]. Available: \url{https://arxiv.org/abs/2605.14473}
\BIBentrySTDinterwordspacing

\bibitem{ye2026seeing}
H.~Ye, S.~Chen, Z.~Zhong, C.~Xiao, H.~Zhang, Y.~Wu, and F.~Shen, ``Seeing
  through the conflict: Transparent knowledge conflict handling in
  retrieval-augmented generation,'' in \emph{Proceedings of the AAAI Conference
  on Artificial Intelligence}, vol.~40, no.~40, 2026, pp. 34\,423--34\,431.

\bibitem{qian2026evidenceforce}
\BIBentryALTinterwordspacing
P.~Qian, S.~Wang, X.~Wang, Y.~Chen, W.~Xu, Q.~Yu, S.~Lin, S.~Zhang, J.~You, and
  X.~Wei, ``Relevant is not warranted: Evidence-force calibration for cited
  {RAG},'' 2026. [Online]. Available: \url{https://arxiv.org/abs/2605.28044}
\BIBentrySTDinterwordspacing

\bibitem{liu2026architecture}
\BIBentryALTinterwordspacing
J.~Liu, J.~Yang, X.~Li, W.~Yan, Y.~Wu, P.~Liang, and M.~Yuan, ``Architecture
  matters more than scale: A comparative study of retrieval and memory
  augmentation for financial {QA} under {SME} compute constraints,'' 2026.
  [Online]. Available: \url{https://arxiv.org/abs/2604.17979}
\BIBentrySTDinterwordspacing

\bibitem{qi2026coverageaware}
\BIBentryALTinterwordspacing
Y.~Qi, Y.~Qi, and T.~Wagh, ``Coverage-aware web crawling for domain-specific
  supplier discovery via a web--knowledge--web pipeline,'' 2026. [Online].
  Available: \url{https://arxiv.org/abs/2602.24262}
\BIBentrySTDinterwordspacing

\bibitem{qi2026correctivehints}
\BIBentryALTinterwordspacing
Y.~Qi, X.~Xu, and Y.~Li, ``When corrective hints hurt: Prompt design in
  reasoner-guided repair of {LLM} overcaution on entailed negations under
  {OWL}~2~{DL},'' 2026. [Online]. Available:
  \url{https://arxiv.org/abs/2604.23398}
\BIBentrySTDinterwordspacing

\bibitem{li2025encoder}
Z.~Li, Z.~Chen, H.~Wen, Z.~Fu, Y.~Hu, and W.~Guan, ``Encoder: Entity mining and
  modification relation binding for composed image retrieval,'' in
  \emph{Proceedings of the AAAI Conference on Artificial Intelligence},
  vol.~39, no.~5, 2025, pp. 5101--5109.

\bibitem{li2025finecir}
Z.~Li, Z.~Fu, Y.~Hu, Z.~Chen, H.~Wen, and L.~Nie, ``{FineCIR}: Explicit parsing
  of fine-grained modification semantics for composed image retrieval,''
  \emph{arXiv preprint arXiv:2503.21309}, 2025.

\bibitem{chen2026optionhedging}
Z.~Chen, M.~Hu, J.~Yi, and W.~Sun, ``Reinforcement learning for option hedging:
  Static implied-volatility fit versus shortfall-aware performance,'' 2026.

\bibitem{zhang2026finsentllm}
Z.~Zhang, R.~Fu, Y.~He, X.~Shen, Y.~Wang, X.~Du, H.~You, K.~Jin, J.~Shi, and
  S.~Fong, ``{FinSentLLM}: Multi-{LLM} and structured semantic signals for
  enhanced financial sentiment forecasting,'' in \emph{ICASSP 2026 -- 2026 IEEE
  International Conference on Acoustics, Speech and Signal Processing
  (ICASSP)}.\hskip 1em plus 0.5em minus 0.4em\relax IEEE, 2026, pp.
  17\,682--17\,686.

\bibitem{qin2025counterfactual}
X.~Qin, S.~Li, Y.~Cai, and L.~Wang, ``Enhancing counterfactual explanations
  with feasibility and diversity,'' in \emph{2025 IEEE International Conference
  on Data Mining Workshops (ICDMW)}.\hskip 1em plus 0.5em minus 0.4em\relax
  IEEE, 2025, pp. 2310--2319.

\bibitem{yang2026llmmentalhealth}
J.~Yang, T.~Liu, Y.~T. Luo, T.~Niu, P.~Pang, A.~Xiang, and Q.~Yang, ``Exploring
  the application boundaries of {LLMs} in mental health: A systematic scoping
  review,'' \emph{Frontiers in Psychology}, vol.~16, p. 1715306, 2026.

\bibitem{cheng2026enhancingfinancialreportquestionanswering}
\BIBentryALTinterwordspacing
Z.~Cheng, L.~Lai, Y.~Liu, K.~Cheng, and X.~Qi, ``Enhancing financial report
  question-answering: A retrieval-augmented generation system with reranking
  analysis,'' 2026. [Online]. Available: \url{https://arxiv.org/abs/2603.16877}
\BIBentrySTDinterwordspacing

\bibitem{cheng2026resolvingrobustnessprecisiontradeofffinancial}
\BIBentryALTinterwordspacing
Z.~Cheng, L.~Lai, and Y.~Liu, ``Resolving the robustness-precision trade-off in
  financial {RAG} through hybrid document-routed retrieval,'' 2026. [Online].
  Available: \url{https://arxiv.org/abs/2603.26815}
\BIBentrySTDinterwordspacing

\bibitem{chen2025autoneuralcodesigningvisionlanguagemodels}
\BIBentryALTinterwordspacing
W.~Chen, L.~Wu, Y.~Hu, Z.~Li, Z.~Cheng, Y.~Qian, L.~Zhu, Z.~Hu, L.~Liang,
  Q.~Tang, Z.~Liu, and H.~Yang, ``{AutoNeural}: Co-designing vision-language
  models for {NPU} inference,'' 2025. [Online]. Available:
  \url{https://arxiv.org/abs/2512.02924}
\BIBentrySTDinterwordspacing

\bibitem{strubell2019energy}
E.~Strubell, A.~Ganesh, and A.~McCallum, ``Energy and policy considerations for
  deep learning in {NLP},'' in \emph{Proceedings of the 57th Annual Meeting of
  the Association for Computational Linguistics}, 2019, pp. 3645--3650.

\bibitem{patterson2021carbon}
D.~Patterson, J.~Gonzalez, Q.~Le, C.~Liang, L.-M. Munguia, D.~Rothchild, D.~So,
  M.~Texier, and J.~Dean, ``Carbon emissions and large neural network
  training,'' \emph{arXiv preprint arXiv:2104.10350}, 2021.

\bibitem{johnson2019faiss}
J.~Johnson, M.~Douze, and H.~J{\'e}gou, ``Billion-scale similarity search with
  {GPUs},'' \emph{IEEE Transactions on Big Data}, vol.~7, no.~3, pp. 535--547,
  2021.

\bibitem{cormack2009rrf}
G.~V. Cormack, C.~L.~A. Clarke, and S.~Buettcher, ``Reciprocal rank fusion
  outperforms condorcet and individual rank learning methods,'' in
  \emph{Proceedings of the 32nd International ACM SIGIR Conference on Research
  and Development in Information Retrieval}.\hskip 1em plus 0.5em minus
  0.4em\relax ACM, 2009, pp. 758--759.

\bibitem{ragminiwikipedia}
{RAG Datasets contributors}, ``{rag-mini-wikipedia}: a small wikipedia-passage
  benchmark for retrieval-augmented generation,'' Hugging Face dataset,
  \url{https://huggingface.co/datasets/rag-datasets/rag-mini-wikipedia}, 2024.

\bibitem{linq2024finder}
{Linq AI Research}, ``{FinDER}: a financial-domain question-answering benchmark
  over {SEC} filings,'' Hugging Face dataset,
  \url{https://huggingface.co/datasets/Linq-AI-Research/FinDER}, 2024.

\bibitem{zheng2023llmjudge}
L.~Zheng, W.-L. Chiang, Y.~Sheng, S.~Zhuang, Z.~Wu, Y.~Zhuang, Z.~Lin, Z.~Li,
  D.~Li, E.~P. Xing, H.~Zhang, J.~E. Gonzalez, and I.~Stoica, ``Judging
  {LLM-as-a-Judge} with {MT-Bench} and {Chatbot Arena},'' in \emph{Advances in
  Neural Information Processing Systems (NeurIPS) Datasets and Benchmarks
  Track}, 2023.

\bibitem{li2025judgesurvey}
D.~Li, B.~Jiang, L.~Huang, A.~Beigi, C.~Zhao, Z.~Tan, A.~Bhattacharjee,
  Y.~Jiang, C.~Chen, T.~Wu \emph{et~al.}, ``From generation to judgment:
  Opportunities and challenges of {LLM-as-a-Judge},'' in \emph{Proceedings of
  the 2025 Conference on Empirical Methods in Natural Language Processing},
  2025, pp. 2757--2791.

\bibitem{lin2026reflectguard}
\BIBentryALTinterwordspacing
L.~Lin, J.~You, Y.~Li, L.~Lin, Y.~Wang, Z.~Zhang, and M.~Zheng,
  ``Reflect-guard: Enhancing {LLM} safeguards against adversarial prompts via
  logical self-reflection,'' 2026. [Online]. Available:
  \url{https://arxiv.org/abs/2605.24834}
\BIBentrySTDinterwordspacing

\bibitem{wang2026visualleakbench}
\BIBentryALTinterwordspacing
Y.~Wang, Y.~Tang, Y.~Qian, and C.~Zhao, ``{VisualLeakBench}: Auditing the
  fragility of large vision-language models against {PII} leakage and social
  engineering,'' 2026. [Online]. Available:
  \url{https://arxiv.org/abs/2603.13385}
\BIBentrySTDinterwordspacing

\bibitem{wang2026safeskills}
\BIBentryALTinterwordspacing
S.~Wang, P.~Qian, Y.~Chen, J.~You, X.~Wang, X.~Jiang, L.~Liu, H.~Yu, and J.~Xu,
  ``When safe skills collide: Measuring compositional risk in agent skill
  ecosystems,'' 2026. [Online]. Available:
  \url{https://arxiv.org/abs/2606.00448}
\BIBentrySTDinterwordspacing

\end{thebibliography}

\end{document}